\documentclass{sigchi}

\usepackage{balance}  
\usepackage{graphics} 
\usepackage{url}      
\usepackage{color}
\usepackage{enumitem}
\usepackage{listings}
\usepackage{pxfonts}
\usepackage{times}    
\usepackage{color}
\usepackage{float}
\usepackage{comment}
\usepackage{tabularx}
\usepackage{amssymb}
\usepackage{booktabs}

\definecolor{codegreen}{rgb}{.38,0.65,.35}
\definecolor{codegray}{rgb}{0.69,0.69,0.69}
\definecolor{codepurple}{rgb}{0.58,0,0.82}
\definecolor{codeblue}{rgb}{0.3,.45,0.6}
\definecolor{codeblack}{rgb}{.08,.08,.08}

\lstdefinestyle{mystyle}{
  commentstyle=\color{codegreen},
  keywordstyle=\color{codeblack},
  numberstyle=\tiny\color{codepurple},
  stringstyle=\color{codegreen},
  basicstyle=\scriptsize\ttfamily\color{codeblack},
  breakatwhitespace=false,         
  breaklines=true,                 
  captionpos=b,                    
  keepspaces=true,                 
  numbers=none,                    
  numbersep=5pt,                  
  showspaces=false,                
  showstringspaces=false,
  showtabs=false,                  
  tabsize=2,
  escapeinside={(*@}{@*)},
}

\definecolor{light-gray}{gray}{0.95}
\definecolor{almost-black}{gray}{0.3}

\makeatletter
\def\url@leostyle{%
  \@ifundefined{selectfont}{\def\UrlFont{\sf}}{\def\UrlFont{\small\bf\ttfamily}}}
\makeatother
\def\pprw{8.5in}
\def\pprh{11in}

\usepackage[pdftex]{hyperref}
\hypersetup{
pdftitle={Empath: Understanding Topic Signals in Large-Scale Text},
pdfauthor={},
pdfkeywords={information extraction, fiction, crowdsourcing, data mining},
bookmarksnumbered,
pdfstartview={FitH},
colorlinks,
citecolor=black,
filecolor=black,
linkcolor=black,
urlcolor=black,
breaklinks=true,
}

\newenvironment{helvetica}{\fontfamily{phv}\selectfont}{\par}

\begin{document}

\title{Empath: Understanding Topic Signals in Large-Scale Text}

\numberofauthors{1}
\author{
\\ \\
  \alignauthor Ethan Fast, Binbin Chen, Michael S. Bernstein\\
   \affaddr{Stanford University}\\
   \email{\{ethan.fast, msb\}@cs.stanford.edu, bchen45@stanford.edu}\\
}

\teaser{
\centering
\includegraphics[width=\textwidth]{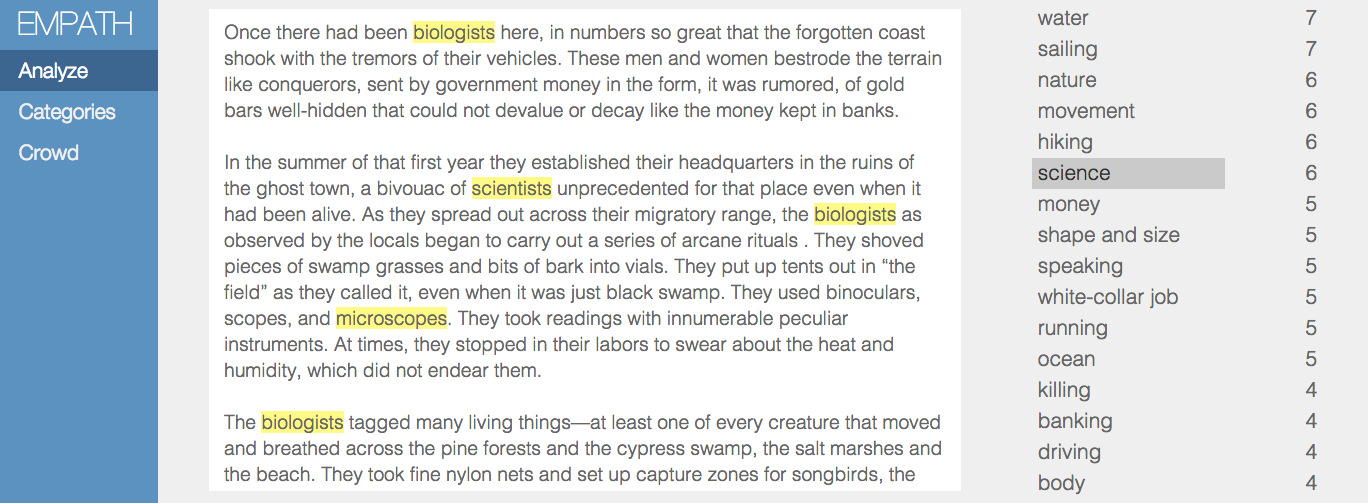}
\caption{Empath analyzes text across 200 gold standard topics and emotions (e.g., childishness or violence), and can generate and validate new lexical categories on demand from a user-generated set of seed terms. The Empath web interface highlights category counts for the current document (right).} 
\label{fig:splash}
}

\maketitle

\begin{abstract}
Human language is colored by a broad range of topics, but existing text analysis tools only focus on a small number of them. We present \textit{Empath}, a tool that can generate and validate new lexical categories on demand from a small set of seed terms (like ``bleed'' and ``punch'' to generate the category \textit{violence}). Empath draws connotations between words and phrases by deep learning a neural embedding across more than 1.8 billion words of modern fiction. Given a small set of seed words that characterize a category, Empath uses its neural embedding to discover new related terms, then validates the category with a crowd-powered filter. 
Empath also analyzes text across 200 built-in, pre-validated categories we have generated from common topics in our web dataset, like \textit{neglect}, \textit{government}, and \textit{social media}. We show that Empath's data-driven, human validated categories are highly correlated (r=0.906) with similar categories in LIWC.
\end{abstract}

\category{H.5.2.}{Information Interfaces and Presentation}{Group and Organization Interfaces}

\section{Introduction}
Language is rich in subtle signals.
The previous sentence, for example, conveys connotations of \textit{wealth} (``rich''), \textit{cleverness} (``subtle''), \textit{communication} (``language'', ``signals''), and \textit{positive sentiment} (``rich''). A growing body of work in human-computer interaction, computational social science and social computing uses tools to identify these signals: for example, detecting emotional contagion in status updates \cite{contagion}, linguistic correlates of deception \cite{deception}, or conversational signs of betrayal \cite{betrayal}. As we gain access to ever larger and more diverse datasets, it becomes important to scale our ability to conduct such analyses with breadth and accuracy. 

\begin{figure*}[!t]
\centering
\includegraphics[width=2.0\columnwidth]{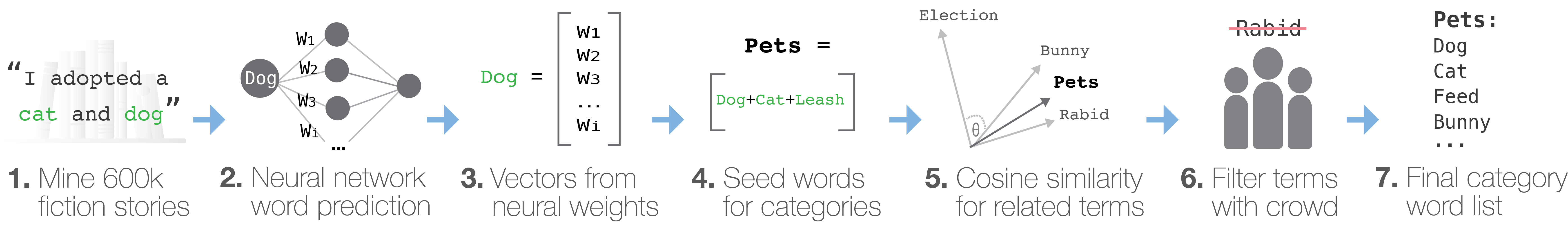}
\caption{Empath learns word embeddings from 1.8 billion words of fiction, makes a vector space from these embeddings that measures the similarity between words, uses seed terms to define and discover new words for each of its categories, and finally filters its categories using crowds.}
\label{fig:overview}
\end{figure*}

High quality lexicons allow us to analyze language at scale and across a broad range of signals. For example, researchers often use LIWC (Linguistic Inquiry and Word Count) to analyze social media posts, counting words in lexical categories like \textit{sadness}, \textit{health}, and \textit{positive emotion} \cite{liwcll}. LIWC offers many advantages: it is fast, easy to interpret, and extensively validated. Researchers can easily inspect and modify the terms in its categories --- word lists that, for example, relate ``scream'' and ``war'' to the emotion \textit{anger}.  But like other popular lexicons, LIWC is small: it has only 40 topical and emotional categories, many of which contain fewer than 100 words. 
Further, many potentially useful categories like \textit{violence} or \textit{social media} don't exist in current lexicons, requiring the ad hoc curation and validation of new gold standard word lists. Other categories may benefit from updating with modern terms like ``paypal'' for \textit{money} or ``selfie'' for \textit{leisure}. 

To address these problems, we present \textit{Empath}: a living lexicon mined from modern text on the web. Empath allows researchers to generate and validate new lexical categories on demand, using a combination of deep learning and crowdsourcing. For example, using the seed terms ``twitter'' and ``facebook,'' we can generate and validate a category for \textit{social media}. Empath also analyzes text across 200 built-in, pre-validated categories drawn from existing knowledge bases and literature on human emotions,
like \textit{neglect} (deprive, refusal), \textit{government} (embassy, democrat), \textit{strength} (tough, forceful), and \textit{technology} (ipad, android). 
Empath combines modern NLP techniques with the benefits of handmade lexicons: its categories are transparent word lists, easily extended and fast. And like LIWC (but unlike other machine learning models), Empath's contents are validated by humans.

While Empath presents an approach that can be trained on any text corpora, in this paper we use 1.8 billion words of modern amateur fiction. 
Why would \emph{fiction} be the right tool to train an externally-valid measure of topical and emotional categories? 
General web text suffers from sparsity when learning categories focused on the human internal states (e.g., \textit{remorse}) or the physical world, for example connecting ``circular'' and ``boxy'' to the topic \textit{shape and size} \cite{kamvar2011we,augur}. On the other hand,  amateur fiction tends to be explicit about both scene-setting and emotion, with a higher density of adjective descriptors (e.g., ``the broken vending machine \textit{perplexed} her.''). Fiction is filled with emotion and description --- it is what gives novels their appeal.

To build Empath, we extend a deep learning skip-gram network to capture words in a neural embedding \cite{word2vec}. This embedding learns associations between words and their context, providing a model of connotation. We can then use similarity comparisons in the resulting vector space to map a vocabulary of 59,690 words onto Empath's 200 categories (and beyond, onto user-defined categories). For example, the word ``self-harming'' shares high cosine similarity with the categories \textit{depressed} and \textit{pain}. Finally, we demonstrate how we can filter these relationships through the crowd to efficiently construct new, human validated dictionaries.

We show how the open-ended nature of Empath's model can replicate and extend classic work in classifying deceptive language \cite{deception}, identifying the patterns of language in movie reviews \cite{movies}, and analyzing mood on twitter \cite{golder-macy}. For example, Empath reveals that 13 emotional categories are elevated in the language of liars, suggesting a novel result that liars tend to use more evocative language. In the movie review dataset, we find positive reviews are more strongly connected with intellectual categories like philosophy, politics, and law. 

Our evaluation validates Empath by comparing its analyses against LIWC, a lexicon of gold standard categories that have been psychometrically validated. We find the correlation between Empath and LIWC across a mixed-corpus dataset is high (r=0.906), and remains high even without the crowd filter (0.90), which suggests Empath's data-driven word counts are very similar to those made by a heavily validated dictionary.
When we instead train Empath on the 100 billion word Google News corpus, its unsupervised model shows less agreement (0.84) with LIWC. In sum, Empath shares high correlation with gold standard lexicons, yet it also offers analyses over a broad and dynamic set of categories.

This paper's contributions include:
\begin{itemize}
\item \textit{Empath}: a text analysis tool that allows users to construct and validate new categories on demand using a few seed terms. It also covers a broad, pre-validated set of 200 emotional and topical categories.
\item An approach to generating and validating word classification dictionaries using a combination of deep learning and microtask crowdsourcing.
\item Results that suggest Empath can generate categories extremely similar to categories that have been hand-tuned and psychometrically validated by humans (average Pearson correlation of 0.906), even without a crowd filter (0.90). 
\end{itemize}

\begin{figure*}[!t]
\centering
\includegraphics[width=2.0\columnwidth]{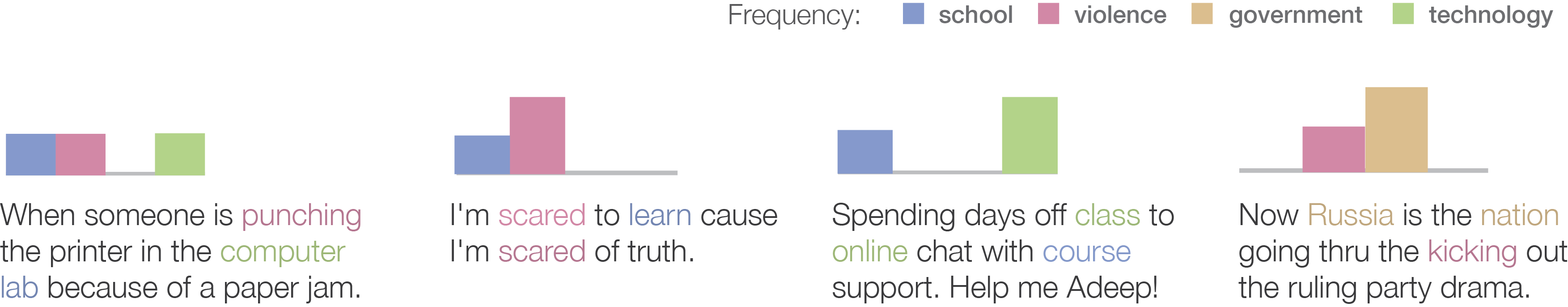}
\caption{Example counts over four example categories on sample tweets from our validation dataset. Words in each category are discovered by unsupervised language modeling, and then validated by crowds.}
\label{fig:overview}
\end{figure*}

\section{Related Work}
Empath inherits from a rich ecosystem of tools and applications for text analysis, and draws on the insights of prior work in data mining and unsupervised language modeling.

\subsection{Extracting Signal from Text}
Text analysis via dictionary categories has a long history in academic research. LIWC, for example, is an extensively validated dictionary that offers a total of 62 syntactic (e.g., present tense verbs, pronouns), topical (e.g., home, work, family) and emotional (e.g., anger, sadness) categories \cite{liwcll}. The General Inquirer (GI) is another human curated dictionary that operates over a broader set of topics than LIWC (e.g., power, weakness), but fewer emotions \cite{gi}. Other tools like EmoLex, ANEW, and SentiWordNet are designed to analyze larger sets of emotional categories \cite{emolex,anew,sentiwordnet}. While Empath's analyses are similarly driven by dictionary-based word counts, Empath operates over a more extensive set of categories, and can generate and validate new categories on demand using unsupervised language modeling.

Work in sentiment analysis, in combination with deep learning, has developed powerful techniques to classify text across positive and negative polarity \cite{socher-rnn}, but has also benefited from simpler, transparent models and rules \cite{vader}. Empath draws on the complementary strengths of these ideas, using the power of unsupervised deep learning to create \textit{human-interpretable} feature sets for the analysis of text. One of Empath's goals is to embed modern NLP techniques in a way that offers the transparency of dictionaries like LIWC. 

\subsection{Applications for Text Analysis}
As social computing and computational social science researchers have gained access to large textual datasets, they have increasingly adopted analyses that cover a wide range of textual signal. For example, researchers have investigated the public's response to major holidays and news events \cite{bollen2009modeling}, how conversational partners mirror each others \cite{kim2012you}, the topical and emotional content of blogs \cite{emolex, kamvar2011we, neviarouskaya2007narrowing}, and whether one person's writing may influence her friends when she posts to social media like Facebook \cite{contagion} or Twitter \cite{twitter-emotion}. Each of these analyses builds a model of the categories that represent their constructs of interest, or uses a word-category dictionary such as LIWC. Through Empath, we aim to empower researchers with the ability to generate and validate these categories. 

Other work in human-computer interaction has relied upon text analysis tools to build new interactive systems. For example, researchers have automatically generated audio transitions for interviews, cued by signals of mood in the transcripts \cite{maneesh-tal}, dynamically generated soundtracks for novels using an emotional lexicon \cite{emolex-music}, or mapped ambiguous natural language onto its visual meaning \cite{vis-lang}. Empath's ability to generate lexical categories on demand potentially enables new interactive systems, cued on nuanced emotional signals like \textit{jealousy}, or diverse topics that fit the new domain.

\subsection{Data Mining and Modeling}
A large body of prior work has investigated unsupervised language modeling. For example, researchers have learned sentiment models from the relationships between words \cite{adj-sem}, classified the polarity of reviews in an unsupervised fashion \cite{unsuper-review}, discovered patterns of narrative in text \cite{nlpnarrative}, and (more recently) used neural networks to model word meanings in a vector space \cite{word2vec, vectorspace}. We borrow from the last of these approaches in constructing of Empath's unsupervised model.

Empath also takes inspiration from techniques for mining human patterns from data. Augur likewise mines fiction, but it does so to learn human activities for interactive systems \cite{augur}. Augur's evaluation indicated that with regard to low-level behaviors such as actions, fiction provides a surprisingly accurate mirror of human behavior. Empath contributes a different perspective, that fiction can be an appropriate tool for learning a breadth of topical and emotional categories, to the benefit of social science. In other research communities, systems have used unsupervised models to capture emergent practice in open source code \cite{codex} or design \cite{webzeitgeist}. In Empath, we adapt these techniques to mine natural language for its relation to emotional and topical categories.

Finally, Empath also benefits from prior work in commonsense knowledge representation. Existing databases of linguistic and commonsense knowledge provide networks of facts that computers should know about the world \cite{conceptnet, wordnet, sentiwordnet}. We draw on some of this knowledge, like the ConceptNet hierarchy, when seeding Empath's categories. Further, Empath itself captures a set of relations on the topical and emotional connotations of words. Some aspects of these connotations may be mineable from social media, if they are of the sort that people are likely to advertise on Twitter \cite{emreactions}. We find that fiction offers a richer source of affective signal.

\section{Empath Applications}
To motivate the opportunities that Empath creates, we first present three example analyses that illustrate its breadth and flexibility. In general, Empath allows researchers to perform text analyses over a broader set of topical and emotional categories than existing tools, and also to create and validate new categories on demand. Following this section, we explain the techniques behind Empath's model in more detail.

\begin{figure}[t!]
\centering
\includegraphics[width=1.0\columnwidth]{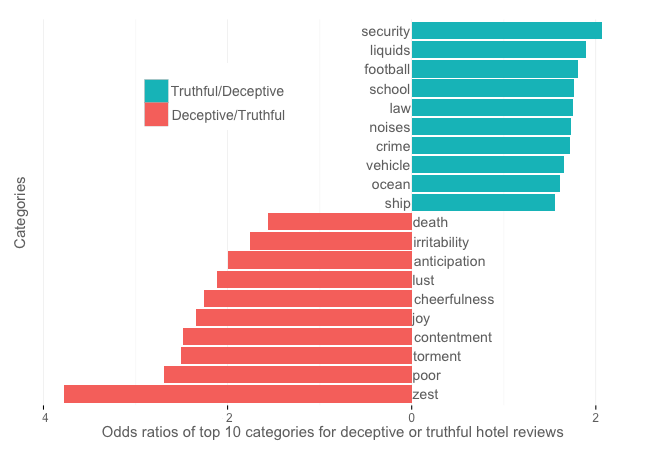}
\caption{Deceptive reviews convey stronger sentiment across both positively and negatively charged categories. In contrast, truthful reviews show a tendency towards more mundane activities and physical objects.}
\label{fig:hotel}
\end{figure}

\subsection{Example 1: Understanding deception in hotel reviews}
What kinds of words accompany our lies? In our first example, we use Empath to analyze a dataset of deceptive hotel reviews reported previously by Ott el al. \cite{deception}. This dataset contains 3200 truthful hotel reviews mined from TripAdvisor.com and deceptive reviews created by workers on Amazon Mechanical Turk, split among positive and negative ratings. The original study found that liars tend to write more imaginatively, use less concrete language, and incorporate less spatial information into their lies.

\subsubsection{Exploring the deception dataset}

We ran Empath's full set of categories over the truthful and deceptive reviews, and produced aggregate statistics for each. Using normalized means of the category counts for each group, we then computed odds ratios and p-values for the categories most likely to appear in deceptive and truthful reviews. All the results we report are significant after a Bonferroni correction ($\alpha = 2.5e^{-5}$).

Our results provide new evidence in support of the Ott et al.\ study, suggesting that deceptive reviews convey stronger sentiment across both positively and negatively charged categories, and tend towards exaggerated language (Figure \ref{fig:hotel}). The liars more often use language that is \textit{tormented} (2.5 odds) or \textit{joyous} (2.3 odds), for example ``it was \textbf{torture} hearing the sounds of the elevator which just would never stop'' or ``I got a \textbf{great} deal and I am so \textbf{happy} that I stayed here.'' The truth-tellers more often discuss concrete ideas and phenomena like the  \textit{ocean} (1.6 odds,), \textit{vehicles} (1.7 odds) or \textit{noises} (1.7 odds), for example ``It seemed like a nice enough place with reasonably close \textbf{beach} access'' or ``they took forever to Valet our \textbf{car}.'' We see a tendency towards more mundane activities among the truth-tellers through categories like \textit{eating} (1.3 odds),  \textit{cleaning} (1.3 odds), or \textit{hygiene} (1.2 odds). ``I ran the \textbf{shower} for ten minutes without ever receiving any hot water.'' For the liars interactions seem to be more evocative, involving  \textit{death} (1.6 odds) or \textit{partying} (1.3 odds). ``The \textbf{party} that keeps you awake will not be your favorite band practicing for their next \textbf{concert}.''

For exploratory research questions, Empath provides a high-level view over many potential categories, some of which a researcher may not have thought to investigate. Lying hotel reviewers, for example, may not have realized they give themselves away by fixating on  \textit{smell} (1.4 odds), ``the room was \textbf{pungent} with what \textbf{smelled} like human excrement'', or their systematic overuse of emotional terms, producing significantly higher odds ratios for 13 of Empath's 32 emotional categories. Truthful reviews, on the other hand, display higher odds ratios for none of Empath's emotional categories.

\subsubsection{Spatial language in lies}
While the original study provided some evidence that liars use less spatially descriptive language, it wasn't able to test the theory directly. Using Empath, we can generate a new set of human validated terms that capture this idea, creating a new \textit{spatial} category. To do so, we tell Empath to seed the category with the terms ``big'', ``small'', and ``circular''. Empath then discovers a series of related terms and uses the crowd to validate them, producing the cluster:

\begin{quote}
\footnotesize
circular, small, big, large, huge, gigantic, tiny, rectangular, rectangle, massive, giant, enormous, smallish, rounded, middle, oval, sized, size, miniature, circle, colossal, center, triangular, shape, boxy, round, shaped, decorative, ...
\normalsize
\end{quote}

When we then add the new \textit{spatial} category to our analysis, we find it favors truthful reviews by 1.2 odds ($p<0.001$). Truth-tellers use more spatial language, for example, ``the room that we originally were in had a \textbf{huge} \textbf{square} cut out of the wall that had exposed pipes, bricks, dirt and dust.'' In aggregate, liars are not as apt in these concrete details. 


\begin{figure}[t!]
\centering
\includegraphics[width=1.0\columnwidth]{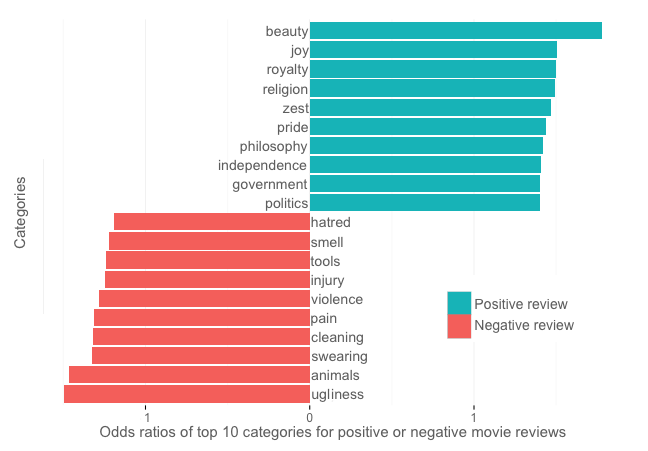}
\caption{Positive reviews are associated with the deeper organizing principals of human life, like politics, philosophy or law. Negative reviews show metaphorical connections to animals, cleaning, and smells.}
\label{fig:movie}
\end{figure}

\subsection{Example 2: Understanding language in movie reviews}

What kinds of movies do reviewers enjoy? What words do reviewers use to praise or pan them? In our second example, we show how Empath can help us discover trends in a dataset of movie reviews collected by Pang et al. \cite{movies}. This dataset contains 2000 movie reviews, divided evenly across positive and negative sentiment.

\subsubsection{Exploring the movie dataset}

The movie review dataset reveals, unsurprisingly, a strong correlation between negative reviews and negatively charged categories (Figure \ref{fig:movie}). For instance, \textit{uglyness} is 1.4 times more likely to appear in a negative review, \textit{swear words} are 1.3  times more likely, and \textit{pain} is 1.3 times more likely. For example, ``oh Bacon glistens when he gets wet all right and looks like a rather \textbf{fatty} side of cheap flank steak'', or ``anything to avoid this \textbf{painful} movie.'' Similarly, positive reviews are associated with more positively charged categories: \textit{beauty} is 1.8 times more likely to appear in a positive review, \textit{joy}  is 1.5 times more likely, and  \textit{pride} is 1.4 times more likely. For example, ``a \textbf{wonderfully} expressive speaking voice full of youthful vigor, and \textbf{gorgeous} singing voice,'' or ``it's the \textbf{triumph} of Secrets \& Lies, then, that it goes beyond gestures of sympathy for the common people.''

Beyond these obviously polarized categories, we find interesting trends in the topics associated with positive and negative reviews. Positive reviews tend to co-occur with the deeper organizing principals of human life, like \textit{politics} (1.4 odds), \textit{philosophy} (1.4 odds), and \textit{law} (1.3 odds) --- possibly indicating the interests of film reviewers. For example, ``Branagh has concentrated on serious issues: \textbf{morality}, \textbf{philosophy} and human elements of the story'', or ``for all of the \textbf{idealism}, it makes me feel good.'' Alternatively, negative reviews adopt what appear to be metaphorical connections to \textit{animals} (1.5 odds), cleaning (1.3 odds), \textit{smell} (1.2 odds) and \textit{alcohol} (1.2 odds). For example, ``no matter how shiny the superficial sheen is, this is still \textbf{trash}, and, like all \textbf{garbage}, it \textbf{stinks}'', or ``a punch-\textbf{drunk} mess of a movie'', or ``no free popcorn coupon can ever restore to us the time we've spent or \textbf{wash} the awful images from our mind.''





The movie dataset also allows us to demonstrate convergent validity between Empath and gold standard tools like LIWC.
For example, are Empath's categories as good as LIWC's for classification? Using five shared emotional categories as features in a logistic regression to predict positive and negative movie reviews, we compare Empath and LIWC under a 10-fold cross-validation t-test that exhibits low Type II error \cite{cross-t}. We find no significant difference between tools ($p=0.43$).

\begin{figure}[t!]
\centering
\includegraphics[width=1.0\columnwidth]{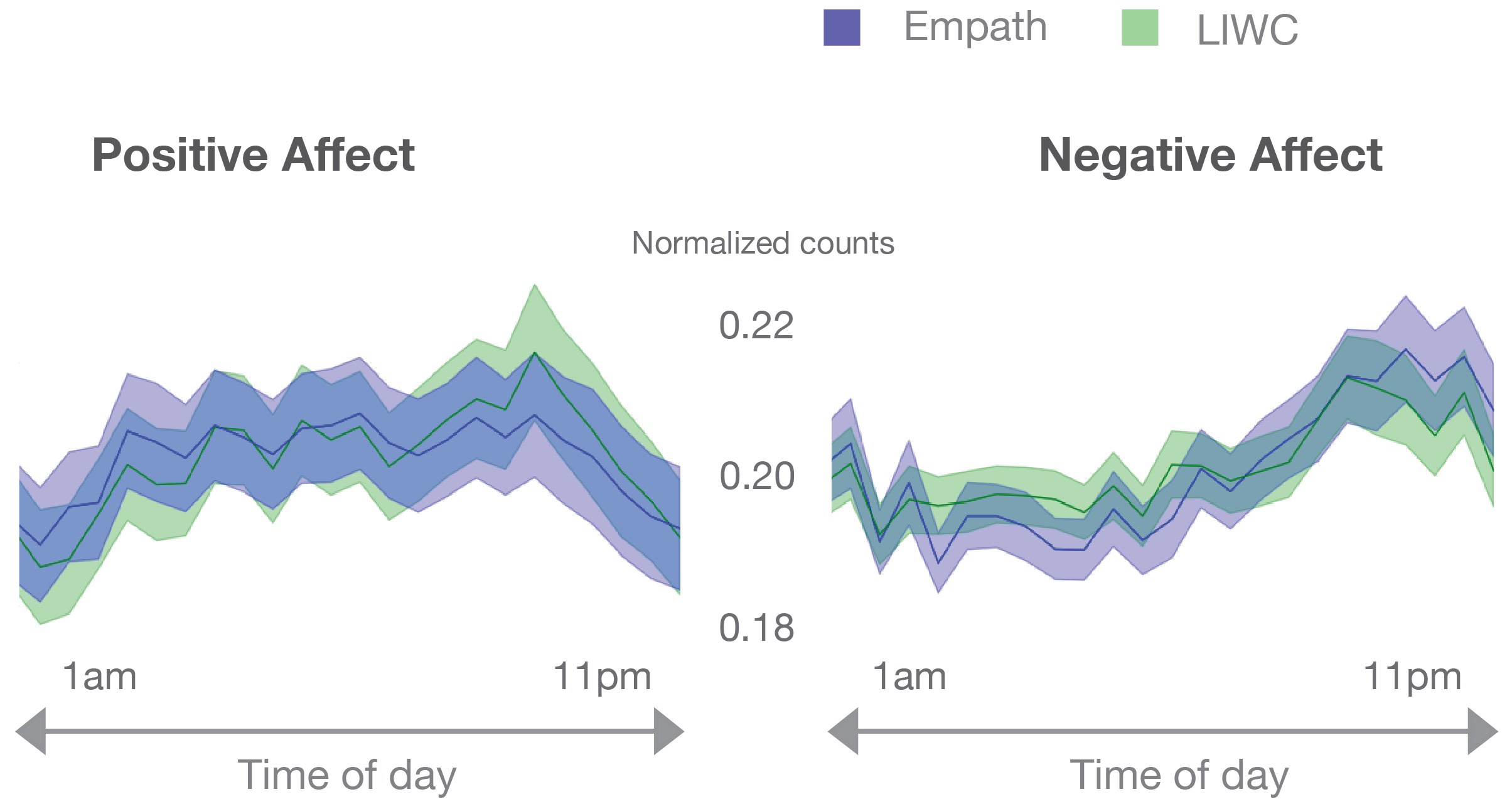}
\caption{We use Empath to replicate the work of Golder and Macy, investigating how mood on Twitter relates to time of day. The signals reported by Empath and LIWC by hour are strongly correlated for positive (r=0.87) and negative (r=0.90) sentiment.}
\label{fig:pos-neg}
\end{figure}

\subsection{Example 3: Mood on Twitter and time of day}

In our final example,
we use Empath to investigate the relationship between mood on twitter and time of day, replicating the work of Golder and Macy \cite{golder-macy}. While the corpus of tweets analyzed by the original paper is not publicly available, we reproduce the paper's findings on a smaller corpus of 591,520 tweets from the PST time-zone, running LIWC on our data as an additional benchmark (Figure \ref{fig:pos-neg}).

The original paper shows a low of negative sentiment in the morning that rises over the rest of the day. We find a similar relationship on our data with both Empath and LIWC: a low in the morning (around 8am), peaking to a high around 11pm. The signals reported by Empath and LIWC over each hour are strongly correlated (r=0.90). Using a 1-way ANOVA to test for changes in mean negative affect by hour, Empath reports a highly significant difference ($F(23,591520)=17.2$, $p<0.001$), as does LIWC ($F=6.8$, $p<0.001$). For positive sentiment, Empath and LIWC again replicate similarly with strong correlation between tools (r=0.87). Both tools once more report highly significant ANOVAs by hour: Empath $F=5.9$, $p<0.001$; LIWC $F=7.3$, $p<0.001$.

\begin{table*}[tb]\scriptsize
\renewcommand{\arraystretch}{1.1}
\begin{helvetica}
  \begin{tabular}{p{6em}@{\hspace{3em}}p{6em}@{\hspace{3em}}p{6em}@{\hspace{3em}}p{6em}@{\hspace{3em}}p{6em}@{\hspace{3em}}p{6em}@{\hspace{3em}}p{6em}@{\hspace{3em}}p{6em}}
  \textbf{social media}  & \textbf{war} & \textbf{violence} & \textbf{technology} & \textbf{fear} & \textbf{pain} & \textbf{hipster} & \textbf{contempt}  \\
  \hline
  facebook   & attack  &  hurt     &   ipad        & horror     &  hurt      & vintage & disdain \\
  instagram & battlefield      &  break            & internet            & paralyze   &  pounding       & trendy & mockery \\
  notification    & soldier  &  bleed  & download             & dread  &  sobbing         & fashion & grudging \\
  selfie  & troop     &    broken      & wireless             & scared    &  gasp & designer & haughty \\
  account     & army  &  scar          & computer   & tremor      &  torment   & artsy & caustic \\
  timeline  & enemy  &  hurting           & email              & despair      &  groan         & 1950s & censure \\
  follower    & civilian     &  injury       & virus             & panic  & stung  & edgy & sneer \\
  \end{tabular}
  \end{helvetica}
\caption{Empath can analyze text across hundreds of data-driven categories. Here we provide a sample of representative terms in 8 sample categories.}
  \label{tbl:cats}
\end{table*}

\section{Empath}

Empath analyzes text across hundreds of topics and emotions. Like LIWC and other dictionary-based tools, it counts category terms in a text document. However, Empath covers a broader set of categories than other tools, and users can generate and validate new categories with a few seed words. 




\subsection{Designing Empath's data-driven categories}
When analyzing textual data, researchers collectively engage with many possible linguistic categories. For example, social scientists study the networks of conversations that surround depression on Twitter \cite{twitter-politics}, psychologists the role of self-presentation in online dating communities \cite{dating}, or digital humanists the role of femininity in greek literature \cite{femininity}.

Empath aims to make possible all of these analyses (and more) through its 200 human validated categories, which cover topics like \textit{violence}, \textit{depression}, or \textit{femininity}. Where do the names of these categories come from? We adopt a data-driven approach using the ConceptNet knowledge base \cite{conceptnet}. The dependency relationships in ConceptNet provide a hierarchy of information and facts that act as a source of category names and seed words for Empath (e.g., war is a form of \textit{conflict}, running is a form of \textit{exercise}). We prefer this approach to a purely manual one as it can potentially scale to thousands of other new categories. 

For example, when a researcher provides ``shirt'' and ``hat'' as seed words, ConceptNet tells us shirts and hats are articles of \textit{clothing}. So, Empath can create and validate a \textit{clothing} category, using ``shirt'' and ``hat'' as seed words:

\begin{quote}
\footnotesize
blazer, vest, sweater, sleeveless, blouse, plaid, tights, undershirt, wearing, jacket, buttondown, longsleeve, skirt, singlet, buttonup, longsleeved, hoody, tanktop, leggings, ...
\normalsize
\end{quote}

Specifically, to generate Empath's category names and seed terms, we selected 200 common dependency relationships in ConceptNet, conditioned on 10,000 common words in our corpus. We then manually refined this list, eliminating redundant or sparse categories. For some categories we added additional seed terms to better represent the concept, resulting in a final set of two to five seed terms for each category.

For emotional analyses, Empath likewise draws upon the hierarchy of  emotions introduced by Parrott \cite{shaver1987emotion}, in which emotions are defined by other emotions. For example, \textit{lust} is defined by ``desire'', ``passion'', and ``infatuation'', so these words become its seed terms in Empath. Similar relationships hold for the other primary and secondary emotions in Parrott's hierarchy, which we use to bootstrap Empath's base set of 32 emotional categories.

While Empath's topical and emotional categories stem from different sources of knowledge, we generate member terms for both kinds of categories in the same way. Given a set of seed terms, Empath learns from a large corpus of text to predict and validate hundreds of similar categorical terms.

\subsection{Learning category terms from a text corpus}
Where do category terms come from? How do we connect a term like ``rampage'' with the category \textit{violent}? Empath takes advantage of recent advances in deep learning to discover category terms in an unsupervised fashion (Figure \ref{fig:overview}).

As we have discussed, each of Empath's categories is defined by seed words (e.g., \textit{lust}: desire, passion; \textit{clothing}: shirt, hat; \textit{social media}: facebook, twitter). Empath's model uses these seed words to generate a candidate set of member terms for its categories, which we validate through paid crowdsourcing. 

Empath generates these category terms by querying a vector space model (VSM) trained by a neural network on a large corpus of text. This VSM allows Empath to examine the similarity between words across many dimensions of meaning. For example, given seed words like ``facebook'' and ``twitter,'  Empath finds related terms like ``pinterest'' and ``selfie.'' 

While Empath can be trained on any text corpus, for the analyses in this paper we use a dataset of modern fiction from Wattpad,\footnote{\url{http://wattpad.com}} a community of amateur writers. This corpus contains more than 1.8 billion words of fiction written by hundreds of thousands of authors. 

\subsubsection{Training a neural vector space model}
To train Empath's model, we adapt the deep learning skip-gram architecture introduced by Mikolov et al. \cite{word2vec}. This is an unsupervised learning task where the basic idea is to teach a neural network to predict co-occurring words in a corpus. For example, the network might learn that ``death'' predicts a nearby occurrence of the word ``carrion,'' but not of ``incest.'' After enough training, the network learns a deep representation of each word that is predictive of its context. We can then borrow these representations, called neural embeddings, to map words onto a vector space. 

More formally, for word $w$ and context $C$ in a network with negative sampling, a skip-gram network will learn weights that maximize the dot product $w\cdot w_c$ and minimize $w\cdot w_n$ for $w_c\in C$ and $w_n$ sampled randomly from the vocabulary. The context $C$ of a word is determined by a sliding window over the document, of a size typically in (0,7).

We train our skip-gram network on the fiction corpus from Wattpad, lemmatizing all words through a preprocessing step. The network uses a hidden layer of 150 neurons (which defines the dimensionality of the embedding space), a sliding window size of five, a minimum word count of thirty (i.e., a word must occur at least thirty times to appear in the training set), negative sampling, and down-sampling of frequent terms. We define and ignore stopwords as words with log-adjusted probability greater than -8, according to the spaCy NLP toolkit.\footnote{\url{http://spacy.io}} These techniques reflect current best practices in deep learning for language models \cite{vectorspace}.

\subsubsection{Building categories with a vector space}
We use the neural embeddings created by our skip-gram network to construct a vector space model (VSM). Similar models trained on neural embeddings, such as word2vec, are well known to enable powerful forms of analogous reasoning (e.g., the vector arithmetic for the terms ``King - Man + Queen'' produces a vector close to ``Woman'') \cite{glove}. In our case, VSMs allow Empath to discover member terms for categories.

VSMs encode concepts as vectors, where each dimension of the vector $v\in\mathbb{R}^{n}$ conveys a feature relevant to the concept. For Empath, each vector $v$ is a word, and each of its dimensions defines the weight of its connection to one of the hidden layer neurons (the neural embeddings). The space is $\mathbb{M}({n\times h})$ where $n$ is the size of our vocabulary (40,000), and $h$ the number of hidden nodes in the network (150).

Empath's VSM selects member terms for its categories (e.g., social media, violence, shame) by using cosine similarity --- a similarity measure over vector spaces --- to find nearby terms in the space. Concretely, cosine similarity is a scaled version of the dot product between two vectors, defined as:
$$cos(\theta) = A\cdot B\;/\;||A||\cdot||B||$$
Using an embedding function $v$ that maps a word to the vector space, we can find the eight terms nearest to $v(\text{depressed})$, by comparing its cosine similarity with all other terms in the space, and selecting the ones that are most similar: 

\begin{quote}
\footnotesize
sad (0.75), heartbroken (0.74), suicidal (0.73), stressed (0.72), self-harming (0.71), mopey (0.70), sick (0.69), moody (0.68) 
\normalsize
\end{quote}

We can also search the vector spaces on \textit{multiple} terms by querying on the vector sum of those terms -- a kind of reasoning by analogy. To generate the query vector for one of Empath's categories, we add the vector corresponding to the name of that category (if it exists in the space), to all the vectors that correspond with its seed terms:
$$query(c,S)=v(c)+\sum_{t\in S}{v(t)}$$
Where $v$ is the embedding function, $c$ is the name a category, and $S$ is a set of seed words that belong to the category.

For example, to generate the terms for \textit{clothing}: \\
\footnotesize
$$query(\text{clothing},\{\text{shirt}, \text{hat}\})=v(\text{clothing})+v(\text{shirt})+v(\text{hat})$$
\normalsize

From a small seed of words, Empath can gather hundreds of terms related to a given category, and then use these terms for textual analysis.

\subsection{Refining categories with crowd validation}
Human-validated categories can ensure that accidental terms do not slip into a lexicon. By filtering Empath's categories through the crowd, we offer the benefits of both modern NLP and human validation: increasing category precision, and more carefully validating category contents. 

To validate each of Empath's categories, we have created a crowdsourcing pipeline on Amazon Mechanical Turk. Specifically, we ask crowdworkers:

\begin{quote}
\footnotesize
For each word, tell us how strongly it relates to the topic. For example, given the topic WAR, the words ``sword'' and``tank'' would be \textit{strongly related}, the word "government" would be \textit{related}, the word ``tactics'' would be \textit{weakly related}, and the words ``house'' and ``desk'' would be \textit{unrelated}.
\normalsize
\end{quote}

Prior work has adopted a similar question and scale \cite{emolex}.

We divide the total number of  words to be filtered across many separate tasks, where each task consists of twenty words to be rated for a given category. For each of these words, workers select a relationship on a four point scale: not related, weakly related, related, and strongly related. We ask three independent workers to complete each task at a cost of \$0.14 per task, resulting in an hourly wage in line with Ethical Guidelines for AMT research \cite{dynamo}. Prior work has shown that three workers are enough for reliable results in labeling tasks, given high quality contributors \cite{get-another-label}. So, if we want to filter a category of 200 words, we would have $200/20=10$ tasks, which must be completed by three workers, at a total cost of $10*3*0.14 = \$4.2$ for this category.

We limit tasks to Masters workers to ensure quality \cite{person-process-crowd} and we aggregate crowdworker feedback by majority vote. If two of three workers believe a word is at least weakly related to the category, then Empath will keep the word, otherwise we remove it from the category. We chose this lower threshold for term inclusion as it showed the highest agreement with LIWC in our benchmarks. On thresholds above weakly related, Turkers tended to discard many more terms, increasing category precision at the cost of much lower recall and hurting overall performance.

\begin{table}[tb]\scriptsize
\renewcommand{\arraystretch}{1.4}
\begin{helvetica}
  \begin{tabular}{p{7em}@{\hspace{1.5em}}p{16.5em}@{\hspace{1.5em}}p{7em}}
  \textbf{Empath Cat.} & \textbf{Words that passed filter} & \textbf{Did not pass}   \\
  \hline
Domestic Work  &   chore, vacuum, scrubbing, laundry & find  \\
Dance              &    ballet, rhythm, jukebox, dj, song  & buds       \\
Aggression               &   lethal, spite, betray, territorial & censure         \\
Attractive               &   alluring, cute, swoon, dreamy, cute & defiantly         \\
Nervousness         &  uneasiness, paranoid, fear, worry & nostalgia \\
Furniture         &  chair, mattress, desk, antique & crate \\
Heroic         &  underdog, gutsy, rescue, underdog  & spoof \\
Exotic         &  aquatic, tourist, colorful, seaside & rural \\
Meeting         &  office, boardroom, presentation  & homework \\
Fashion         &  stylist, shoe, tailor, salon, trendy & yoga \\

\end{tabular}
  \end{helvetica}
\caption{ Crowd workers found 95\% of the words generated by Empath's unsupervised model to be related to its categories. However, machine learning is not perfect, and some unrelated terms slipped through  (``Did not pass'' above), which the crowd then removed.}
  \label{tbl:crowd-filter}
\end{table}

\subsubsection{Contents and Efficiency}
Prior work suggests that category construction is subjective, often resulting in low agreement among humans \cite{liwc-valid,emolex}. How well do human workers agree with Empath? On average over 200 categories, workers rated 96\% of the words generated by Empath as related to its categories, and agreed among themselves (voting unanimously with unrelated or related scores) at a rate of 81\%. We provide a sample of terms accepted and rejected by the crowd in Table \ref{tbl:crowd-filter}. 

An acceptance score of 96\% allows us to efficiently collect validated words for Empath's categories. Prior work in human-validated category construction has typically relied upon less efficient approaches, for example using crowd workers to annotate the 10,000 most common dictionary words with scores over all categories in question \cite{emolex}. Empath's unsupervised accuracy allows us to validate the same size categories with far fewer crowdsourcing tasks.
At scale, the cost of crowdsourcing new lexicons is expensive. To validate 200 categories naively over 5,000 words would cost \$21,000, assuming 14 cents a task. We have validated Empath's 200 categories (with a vocabulary of more than 10,000 words) at a total cost of \$840.

Our experience confirms the findings of prior work that category construction is somewhat subjective. Not all of the words rejected by majority vote are necessarily unrelated to a category, and in fact 36\% of them had a worker cast a minority vote for relevance.

\subsection{Empath API and web service}
Finally, to help researchers analyze text over new kinds of categories, we have released Empath as a web service and open source library. The web service\footnote{\url{http://empath.stanford.edu}} allows users to analyze documents across Empath's built-in categories (Figure \ref{fig:splash}), generate new unsupervised categories, and request new categories be validated using our crowdsourcing pipeline. The open source library\footnote{\url{https://github.com/Ejhfast/empath}} is written in Python and similarly returns document counts across Empath's built-in validated categories.

\section{Evaluation}

Can fiction teach computers about the emotional and topical connotations of words? Here we evaluate Empath's crowd filtered and  unsupervised predictions against similar gold standard categories in LIWC.

\subsection{Comparing Empath and LIWC}
The broad reach of our dataset allows Empath to classify documents among a large number of categories. But how accurate are these categorical associations? Human inspection and crowd filtering of Empath's categories (Table \ref{tbl:crowd-filter}) provide some evidence, but ideally we would like to answer this question in a more quantitative way.

Fortunately, LIWC has been extensively validated by researchers \cite{liwcll}, so we can use it to benchmark Empath's predictions across the categories that they share in common. If we can demonstrate that Empath provides very similar results across these categories, this would suggest that Empath's predictions are close to achieving gold standard accuracy. 

Here we compare the predictions of Empath and LIWC over 12 shared categories: \textit{sadness}, \textit{anger}, \textit{positive emotion}, \textit{negative emotion}, \textit{sexual}, \textit{money}, \textit{death}, \textit{achievement}, \textit{home}, \textit{religion}, \textit{work}, and \textit{health}. 

\subsubsection{Method}
To compare all tools, we created a mixed textual dataset evenly divided among tweets \cite{emolex-tweets}, StackExchange opinions \cite{politeness},  movie reviews \cite{movies},  hotel reviews \cite{deception}, and chapters sampled from four classic novels on Project Gutenberg (David Copperfield, Moby Dick, Anna Karenina, and The Count of Monte Cristo) \cite{gutenberg}. This mixed corpus contains more than 2 million words in total across 4500 individual documents. 

Next we selected two parameters for Empath: the minimum cosine similarity for category inclusion and the seed words for each category (we fixed the size of each category at a maximum of 200 words). To choose these parameters, we divided our mixed text dataset into a training corpus of 900 documents and a test corpus of 3500 documents. We selected up to five seed words that best approximated each LIWC category, and found that a minimum cosine similarity of 0.5 offered the best performance. We created crowd filtered versions of these categories as described in the previous section.

We then ran all tools over the documents in the test corpus, recorded their category word counts, then used these counts to compute Pearson correlations between all shared categories, as well as aggregate overall correlations. Pearson's \textit{r} measures the linear correlation between two variables, and returns a value between (-1,1), where 1 is total positive correlation, 0 is no correlation, and −1 is total negative correlation. In our experiment, these correlations speak to how well one tool approximates another. 

To anchor this analysis, we collected benchmark Pearson correlations against LIWC for GI and EmoLex (two existing human validated lexicons). We found a benchmark correlation of 0.876 between GI and LIWC over \textit{positive emotion}, \textit{negative emotion}, \textit{religion}, \textit{work}, and \textit{achievement}, and a correlation of 0.899 between EmoLex and LIWC over \textit{positive emotion}, \textit{negative emotion}, \textit{anger},  and \textit{sadness}. While EmoLex and GI are commonly regarded as gold standards, they correlate imperfectly with LIWC. We take this as evidence that gold standard lexicons can disagree: if Empath approximates their performance against LIWC, it agrees with LIWC as well as other carefully-validated dictionaries agree with LIWC.

Finally, to test the importance of choosing seed terms, we re-ran our evaluation while permuting the seed words in Empath's categories. Over one trial, we dropped one seed term from each category. Over another, we replaced one term from each category with a similar alternative (e.g., ``church'' to ``chapel'', or ``kill'' to ``murder''). 

\begin{table}[tb]\scriptsize
\renewcommand{\arraystretch}{1.3}
\begin{helvetica}
  \begin{tabular}{p{6em}@{\hspace{1.5em}}p{5em}@{\hspace{1.5em}}p{7em}@{\hspace{2em}}p{4em}@{\hspace{1.5em}}p{3em}}
  \textbf{LIWC Cat.} & \textbf{Empath} & \textbf{Empath+Crowd} & \textbf{Emolex} &  \textbf{GI}  \\
  \hline
Positive                &   0.944 & 0.950   &    0.955    &    0.971 \\
Negative              &    0.941 & 0.936   &    0.954    &    0.945 \\
Sadness               &     0.890 & 0.907   &    0.852    &     \\
Anger                    &    0.889  & 0.894  &    0.837    &     \\
Achievement      &      0.915 &  0.903      &        &    0.817 \\
Religion               &      0.893 & 0.908        &        & 0.902     \\
Work                   &   0.859 & 0.820     &       &    0.745 \\
Home                    &     0.919 & 0.941            &       &    \\
Money                 &       0.902 & 0.878        &       &     \\
Health                 &       0.866 & 0.898       &        &     \\
Sex                 &       0.928 & 0.935       &        &     \\
Death                   &  0.856 & 0.901      &      &    \\
\hline
Average                   &     0.900 & 0.906   &   0.899    &    0.876 \\
\end{tabular}
  \end{helvetica}
\caption{We compared the classifications of LIWC, EmoLex and Empath across thirteen categories, finding strong correlation between tools. The first column represents comparisons between Empath's unsupervised model against LIWC, the second after crowd filtering against LIWC, the third between EmoLex and LIWC, and the fourth between the General Inquirer and LIWC.}
  \label{tbl:comp}
\end{table}

\begin{figure*}[t!]
\centering
\includegraphics[width=2.0\columnwidth]{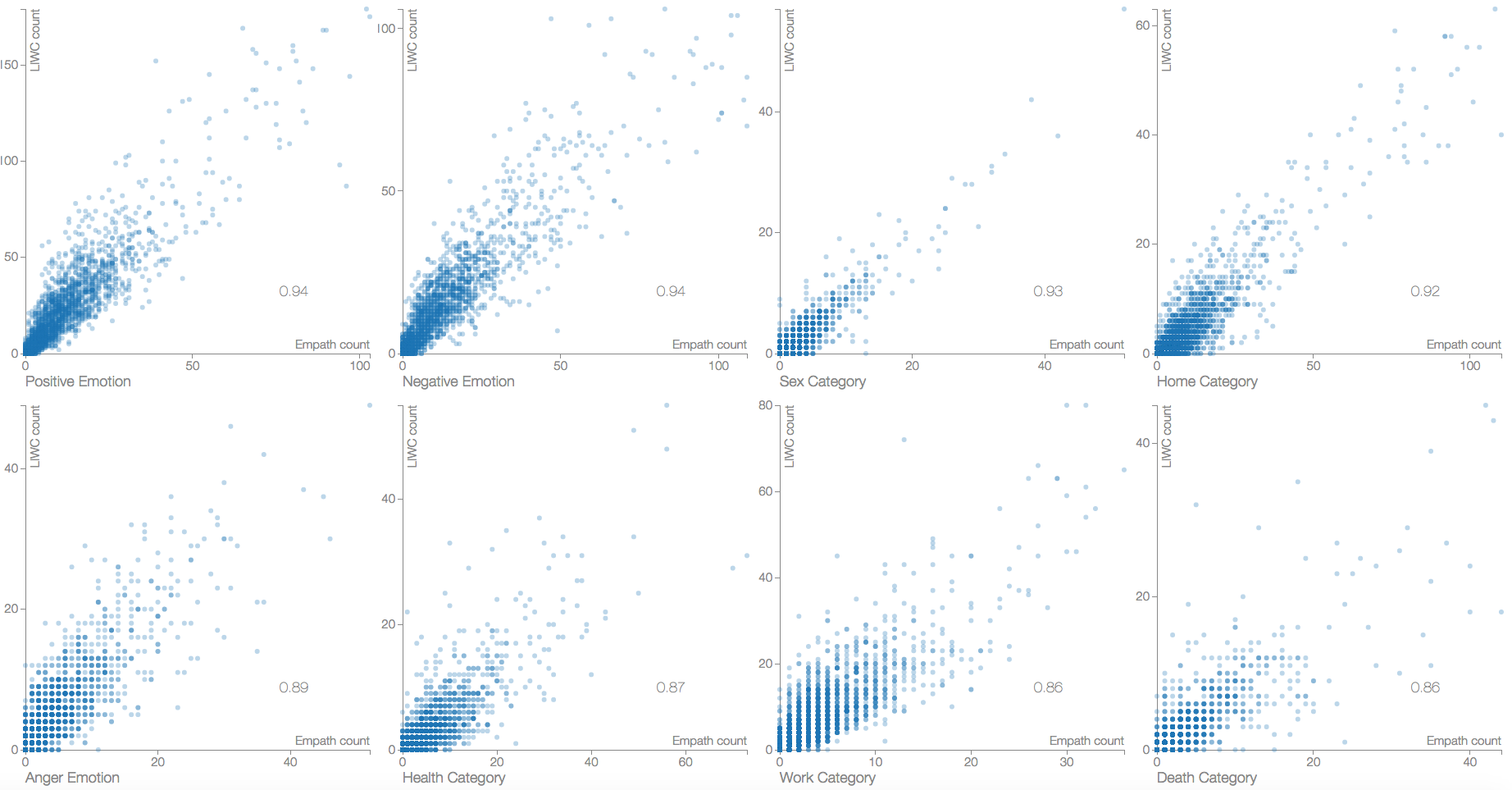}
\caption{Empath categories strongly agreed with LIWC, at an average Pearson correlation of 0.90. Here we plot Empath's best and worst correlations with LIWC. Each dot in the plot corresponds to one document. Empath's counts are graphed on the x-axis, LIWC's on the y-axis.}
\label{fig:scatter}
\end{figure*}

\subsubsection{Results}
Empath shares overall average Pearson correlations of 0.90 (unsupervised) and 0.906 (crowd) with LIWC (Table \ref{tbl:comp}). Over the emotional categories, Empath and LIWC agree at correlations of 0.884 (unsupervised) and 0.90 (crowd), comparing favorably with EmoLex's correlation of 0.899. Over GI's benchmark categories, Empath reports 0.893 (unsupervised) and 0.91 (crowd) correlations against LIWC, stronger performance than GI (0.876). On average, adding a crowd filter to Empath improves its correlations with LIWC by 0.006. We plot Empath's best and worst category correlations with LIWC in Figure \ref{fig:scatter}. These scores indicate that Empath and LIWC are strongly correlated -- similar to the correlation between LIWC and other published and validated tools.

In permuting Empath's seed terms, we found it retained high unsupervised agreement with LIWC (between 0.82 and 0.88). The correlation between tools was most strongly affected when we dropped seeds that added a unique meaning to a category. For example, \textit{death} is seeded with the words ``bury'', ``coffin'', ``kill'', and ``corpse.'' When we removed ``kill'' from the \textit{death}'s seed list, Empath lost the adversarial aspects of death (embodied in words like ``war'', ``execute'', or ``murder'') and fell to 0.82 correlation with LIWC for that category. Removing \textit{death}'s other seed words did not have nearly so strong an affect. On the other hand, replacing seeds with alternative forms or synonyms (e.g., ``hate'' to ``hatred'', or ``kill'' to ``murder'') usually had little impact on Empath's correlations with LIWC.


\section{Discussion}

Empath demonstrates an approach that crosses traditional text analysis metaphors with advances in deep learning. Here we discuss our results and the limitations of our approach.

\subsection{The role of human validation}
While adding a crowd filter to Empath improves its overall correlations with LIWC, the improvement is not statistically significant. Even more surprisingly, the crowd does not always improve agreement at the level of individual categories. For example, across the categories \textit{negative emotion}, \textit{achievement}, and \textit{work}, the crowd filter slightly decreases Empath's agreement with LIWC. When we inspected the output of the crowd filtering step to determine what had caused this effect, we found in a small number of cases in which the crowd was overzealous. For example, the word ``semester'' appears in LIWC's \textit{work} category, but the crowd removed it from Empath. Should ``semester'' be in a \textit{work} category? This disagreement highlights the inherent ambiguity of constructing lexicons. In our case, when the crowd filters out a common word shared by LIWC (like ``semester''), this causes overall agreement across the corpus to decrease (through additional false negatives), despite the appropriate removal of many other less common words. 

As we see in our results, this scenario does not happen often, and when it does happen the effect size is small. 
We suggest that crowd validation offers the qualitative benefit of removing false positives from analyses, while on the whole performing almost identically to (and usually slightly better than) the unfiltered version of Empath. 

\subsection{Data-driven: who is actually driving?}
Empath, like any data-driven system, is ultimately at the mercy of its data -- garbage in, garbage out. 
While fiction allows Empath to learn an approximation of the gold-standard categories that define tools like LIWC, its data-driven reasoning may succeed less well on corner cases of analysis and connotation. Just because fictional characters often pull guns out of gloveboxes, for example, does not mean the two should be strongly connected in Empath's categories.

Contrary to this critique, we have found that fiction is a useful training dataset for Empath given its abundance of concrete descriptors and emotional terms.  
When we replaced the word embeddings learned by our model with alternative embeddings trained on  Google News \cite{word2vec}, we found its average unsupervised correlation with LIWC decreased to 0.84. The Google News embeddings performed better after significance testing on only one category, \textit{death} (0.91), and much worse on several of the others, including \textit{religion} (0.78) and \textit{work} (0.69).  This may speak to the limited influence of fiction bias. Fiction may suffer from the overly fanciful plot events and motifs that surround death (e.g. suffocation, torture), but it captures more relevant words around most categories. 

\subsection{Limitations}
Empath's design decisions suggest a set of limitations, many of which we hope to address in future work. 

First, while Empath reports high Pearson correlations with LIWC's categories, it is possible that other more qualitative properties are important to lexical categories. Two lexicons can be statistically similar on the basis of word counts, and yet one might be easier to interpret than the other, offer more representative words, or present fewer false positives or negatives. At a higher level, the number and kinds of categories available in Empath present a related concern. We created these categories in a data-driven manner. Do they offer the right balance and breadth of topics? We have not evaluated Empath over these more qualitative aspects of usability.

Second, we have not tested how well Empath's categories generalize beyond the core set it shares with LIWC. Do these new categories perform as well in practice? While Empath's categories are all generated and validated in the same way, we have seen though our evaluation that choice of seed words can be important.  What makes for a good set of seed terms? And how do we best discover them? In future work, we hope to investigate these questions more closely. 

Finally, while fiction provides a powerful model for generating lexical categories, we have also seen that, for certain topics (e.g. \textit{death} in Google News), other corpora may have even greater potential. Could different datasets be targeted at specific categories? Mining an online fashion forum, for example, might allow Empath to learn a more comprehensive sense of \textit{style}, or Hacker News might give it a more nuanced view of \textit{technology} and \textit{startups}. We see potential for training Empath on other text beyond fiction.

\subsection{Statistical false positives}
Social science aims to avoid Type I errors --- false claims that statistically appear to be true. Because Empath expands the number of categories available for analysis, it is important to consider the risk of a scientist analyzing so many categories that one of them, through sheer randomness, appears to be elevated in the text. In this paper, we used Bonferroni correction to handle the issue, but there are more mature methods available. For example, Holm's method and FDR are often used in statistical genomics to test thousands of hypotheses. In the case of regression analysis, it is likewise important not to do so-called ``garbage can regressions'' that include every possible predictor. In this case, models that penalize complexity (e.g., non-zero coefficients) are most appropriate, for example LASSO or logistic regression with an L1 penalty.

\section{Conclusion}

Empath aims to combine modern NLP techniques with the transparency of dictionaries like LIWC. In doing so, it provides both broader and deeper forms of text analysis than existing tools.
In breadth, Empath offers hundreds of pre-defined lenses through which researchers can analyze text. In depth, its user-defined categories provide a flexible means by which researchers can ask domain-specific questions. These questions are ever changing, as is our use of language. Empath is a living lexicon -- able to keep up with each.

\section{Acknowledgments}
Special thanks to our reviewers and colleagues at Stanford for their helpful feedback.
This work is supported by a NSF Graduate Fellowship and a NIH and Stanford Medical Scientist Training Grant.





\bibliographystyle{acm-sigchi}
\bibliography{ref}

\end{document}